# Automatic Firewall rules generator for Anomaly Detection Systems with Apriori algorithm

Ehsan Saboori
K.N Toosi University of Technology
Tehran, Iran
ehsansaboori@sina.kntu.ac.ir

Shafigh Parsazad, Yasaman Sanatkhani
Ferdowsi University, University of East London
Mashhad, Iran
London, UK
Shafigh.Parsazad@stu-mail.um.ac.ir

*Abstract*— Network intrusion detection systems have become a crucial issue for computer systems security infrastructures. Different methods and algorithms are developed and proposed in recent years to improve intrusion detection systems. The most important issue in current systems is that they are poor at detecting novel anomaly attacks. These kinds of attacks refer to any action that significantly deviates from the normal behaviour which is considered intrusion. This paper proposed a model to improve this problem based on data mining techniques. Apriori algorithm is used to predict novel attacks and generate real-time rules for firewall. Apriori algorithm extracts interesting correlation relationships among large set of data items. This paper illustrates how to use Apriori algorithm in intrusion detection systems to cerate a automatic firewall rules generator to detect novel anomaly attack. Apriori is the best-known algorithm to mine association rules. This is an innovative way to find association rules on large scale.

## I. INTRODUCTION

An intrusion is defined as any set of actions that attempt to compromise the integrity, confidentiality or availability of a resource. Intrusion detection is classified into two types: misuse intrusion detection and anomaly intrusion detection [1].

Misuse detection is based on known attack actions. In this method features are extracted from known intrusions and rules are pre-defined. The important disadvantage of this method is the novel or unknown attacks that cannot be detected.

Anomaly detection is based on the normal behaviour of a subject; any action that significantly deviates from the normal behaviour is considered intrusion [4]. Sometimes the training audit data does not include intrusion data. One problem with anomaly detection is that it is likely to raise many false alarms.

## II. INTRUSION DETECTION SYSTEMS

Intrusion Detection System (IDS) is a system that recognizes, in real time, attacks and threats on a network and takes corrective action to prevent those attacks and threats. An intrusion detection system is used to detect malicious activities that can compromise the security of the system. There are three key elements necessary in each Intrusion detection system. The first element is Resources to be protected. The second element is the definition of the legitimate action on the resources. Intrusion detection systems must know which action is legal or illegal in order to take correct action to prevent probable attacks. The third element is the efficient methods that act as real-time activities against the models and report probable "intrusive" activities. There are several ways to categorize intrusion detection systems:
- Misuse detection vs. Anomaly detection
- Network-based vs. Host-based systems
- Passive systems vs. Reactive systems

These categories will be explained more in the following paragraphs.

### A. Misuse detection vs. Anomaly detection

In misuse detection, the intrusion detection system analyses the recorded information and compares it to large databases of attack signatures. These methods Record the specific patterns of intrusions as a large rule database, and then monitor activities and pattern matching. Every event that is matched with rules in database is reported as intrusion like a virus detection system. The main problem in misuse detection approach is known intrusion patterns that have to be hand-coded and is unable to detect any novel intrusions.

In anomaly detection, the intrusion detection system establishes the normal behaviour profiles then observes and compares current activities with the normal profiles. If any deviation activities occurred, the system reports it as intrusions. The main problem in anomaly detection is selecting the right set of system features which are ad hoc and based on experience and unable to capture sequential interrelation between events. High false-alarm and limited by training data are other problems and issues in anomaly detection systems.

TABLE 1. Misuse detection vs. Anomaly detection

|  | Advantage | Disadvantage |
|---|---|---|
| Misuse Detection | Accurately and generate much fewer false alarm | Cannot detect novel attacks and threats |
| Anomaly Detection | Is able to detect unknown attacks based on audit | High false-alarm and limited by training data. |

### B. Network-based vs. Host-based systems

A network-based system (NIDS), analyses network packets that are captured on a network. The NIDS can detect malicious packets received on a network. Snort is the most popular network-based Intrusion detection software that performs protocol analysis, content searching/matching and can be used to detect a variety of attacks. Snort is used in proposed method to record a large number of activities for Association rules training dataset. several paradigms have been used to develop diverse NIDS approaches (a detailed analysis of related work in this area can be found for instance in [5]): Expert Systems [6], Finite Automatons [7], Rule Induction Systems [8], Neural Networks [1], Intent Specification Languages [9], Genetic Algorithms [10], Fuzzy Logic [11], Support Vector Machines [1], Intelligent Agent Systems [12] or Data-Mining-based approaches [13]. Still, none of them tries to combine anomaly and misuse detection and, fail when applied to either well-known or zero-day attacks. There is one exception in [14], but the analysis of network packets is too superficial (only headers) to yield any good results in real life.

In a host-based system, the IDS examines the activity on each individual computer or host. A host-based IDS analyses several areas to determine malicious activity inside the network or intrusion from the outside. Each host-based IDS has several types of log files for network, firewall, etc.…, and compares the log files against the database of regular rules and signatures to detect known attacks and threats. Host Intrusion Detection Systems are run on individual hosts or devices on the network. A HIDS monitors only the inbound and outbound packets from the device and will alert the user or administrator if suspicious activity is detected. Host-based systems can also verify the data integrity of important and executable files. It checks a database of sensitive files (and any files added by the administrator) and creates a checksum of each file with a message-file digest utility such as md5sum (128-bit algorithm) or sha1sum (160-bit algorithm). The host-based IDS then stores the sums in a plain text file and periodically compares the file checksums against the values in the text file. If any of the file checksums do not match, the IDS alert the administrator by email or cellular pager. This paper proposed a method to generate real-time firewall rules by using Snort and Apriori algorithm. So this method can be classified as host-based intrusion detection systems.

*C. Passive systems vs. Reactive systems*

In a passive system, the intrusion detection system (IDS) sensor detects a potential security breach, logs the information and signals an alert on the console and/or owner. In a reactive system, also known as an intrusion prevention system (IPS), the IPS responds to the suspicious activity by resetting the connection or by reprogramming the firewall to block network traffic from the suspected malicious source. This can happen automatically or at the command of an operator.

III. DATA MINING TECHNIQUES

Data mining, sometimes called data or knowledge discovery, is the process of analysing data from different perspectives and summarizing it into useful information. This Information can be used to increase system efficiency. Technically, data mining is the process of finding correlations or patterns among dozens of fields in large relational databases. There are several steps for mining data. Understanding the application domain, data preparation, data mining, interpretation, and utilizing the discovered knowledge, after all applying specific algorithms to extract patterns from data. Many algorithms are proposed in data mining approach and implemented but some algorithms are relevant to be used in intrusion detection systems such as:

- *Classification:* maps a data item into one of several pre-defined categories.
- *Link analysis:* determines relations between fields in the database.
- *Sequence analysis:* models sequence patterns.

Data mining approaches for intrusion detection were first implemented in mining audit data for automated models for intrusion detection [2, 3]. Association rules are one of data mining techniques.

IV. ASSOCIATION RULES

Association rule mining extracts interesting correlation relationships among large set of data items. The goal of these techniques is to detect relationships or associations between specific values of categorical variables in large data sets. This is a common task in many data mining projects. These techniques enable analysts and researchers to uncover hidden patterns in large datasets.

Association rules provide information in the form of if-then statements. These rules are computed from the dataset. Unlike logic if-then statement, association rules are probabilistic in nature. An association rules has two numbers that expresses the degree of uncertainty about the rule. These numbers refer to antecedent and consequent. Antecedent refers to the "if" part of rule and consequent refers to the "then" part of rule. Antecedent and consequent are sets of items called itemsets that do not have any items in common.

The first number is called the support for the rule. The support is simply the number of transactions that include all items in the antecedent and consequent parts of the rule. The support is sometimes expressed as a percentage of the total number of records in the database.

The other number is known as the confidence of the rule. Confidence is the ratio of the number of transactions that includes all items in the consequent as well as the antecedent.

Formal model for association rules are shown below:

$I = \{i_1, i_2, \dots, i_m\}$ : Set of itmes
$D$: database of transactions
$T \in D$: a transaction . $T \subseteq I$
$TID$ :unique identifier, associated with each $T$
$X$ : a subset of $I$
$T$ contains $X$ if $X \subseteq T$
Association rule : $X \Rightarrow Y$ here $X \subset I$, $Y \subset I$ and $X \cap Y = \emptyset$
$Supp(X \cup Y) = $ number of transactions in $D$ contain $(X \cup Y)$
$$conf(X \Rightarrow Y) = \frac{supp(X \cup Y)}{supp(X)}$$

Apriori is the best-known algorithm to mine association rules. This algorithm was developed by Agrawal and Srikant in 1994. This is an innovative way to find association rules on large scale, allowing implication outcomes that consist of more than one item.

## V. APRIORI ALGORITHM

Association rules Find frequent itemsets whose occurrences exceed a predefined minimum support threshold and deriving association rules from those frequent itemsets. These two sub problems are solved iteratively until no more new rules emerge. Minimum support threshold must be defined by user and initial transactional database. This algorithm uses knowledge from previous iteration phase to produce frequent itemsets.

For Creating frequent sets let's define :
  $C_k$ as a candidate itemset of size $k$
  $L_k$ as a frequent itemset of size $k$
Main steps of iteration are:
1. Find frequent set $L_{k-1}$
2. Join step: $C_k$ is generated by joining $L_{k-1}$ with itself (Cartesian product $L_{k-1} \times L_{k-1}$)
3. Prune step (Apriori property): Any ($k-1$) Size itemsets that is not frequent cannot be a subset of a frequent $k$ size itemsets, hence should be removed.
4. Frequent set $L_k$ has been achieved

This algorithm uses breadth-first search and a hash tree structure to make candidate itemsets efficient, and then the frequency occurrence for each candidate itemsets will be counted. Those candidate itemsets that have higher frequency than minimum support threshold are qualified to be frequent itemsets. The pseudocode of algorithm is shown below.

```
Apriori(T, ε)
  L₁ ← { large 1-itemsets that appear in more than ε transactions }
  k ← 2
  while L_{k-1} ≠ ∅
    C_k ← Generate(L_{k-1})
    for transactions t ∈ T
      C_t ← Subset(C_k, t)
      for candidates c ∈ C_t
        count[c] ← count[c] + 1
    L_k ← {c ∈ C_k | count[c] ≥ ε}
    k ← k + 1
  return ⋃_k L_k
```

Figure 1. Apriori pseudo code

Apriori algorithm was used in the proposed method to improve anomaly detection systems.

## VI. MOTIVATION FOR USING APRIORI ALGORITHM IN INTRUSION DETECTION

Any intrusive or normal activities leave evidence in audit data that large databases can be recorded from log data. Also from the data-centric point of view, intrusion detection is a data analysis processes, therefore it is an interesting method for using data mining techniques in intrusion detection systems. Audit data can be easily formatted into a database table and program executions and user activities have frequent correlation among system features, therefore association rule is the best choice to be used in this approach. Audit Dataset of user behaviour is too large, so using Apriori algorithm is better than using association rules. The most important advantage is that incremental updating of the rule set is easy. Apriori algorithm creates a model to generate rules for firewall.

## VII. THE PROPOSED METHOD

Snort is the most popular network-based Intrusion detection software that performs protocol analysis, content searching, matching which includes using rule sets and can be used to detect a variety of attacks. Several rule sets are available for being used, including those officially approved by the Sourcefire Vulnerability Research Team (VRT) and those contributed by other communities [15], [16]. Snort supports a simple rule language that matches against network packets, generating alerts or log messages. Snort is used in proposed method to record and log a large number of user's activities such as URLs, visited sites, ports, etc for Apriori training dataset. This dataset is used to create a model to compute the probability of raising the alarm by the current activates. If this probability exceeds defined threshold, this will generate a rule to be added to firewall in order to block it in the future. Figure 2 is illustrated the scenario of method.

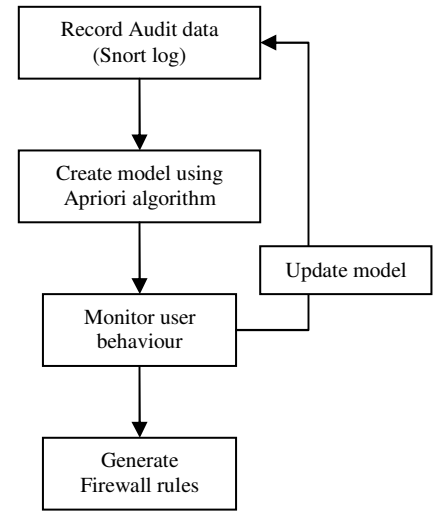

Figure 2. The Scenario of the model

In order to train the proposed model, we used snort to collect information and activities of 10 people in period of 2 weeks. The collected information contains IP Addresses, Transport Layer protocols such as Transmission Control Protocol (TCP) and User Datagram Protocol (UDP) which specify a source and destination port number in their packet headers. The process associates with a particular port (known as binding) to send and receive data, it means that it will listen for incoming packets whose destination port number and IP destination address match that port, or send outgoing packets whose source port number is set to that port. The TCP/IP software on the computer will receive the packets from the driver software and copy them to the processes address space. Port numbers are contained in the transport protocol packet header, and they can be readily interpreted not only by the sending and receiving computers, but also by other components of the networking infrastructure. In particular, firewalls are commonly configured to differentiate between packets depending on their source or destination port numbers. Port forwarding is an example application of this. Port numbers can occasionally be seen in the Uniform Resource Locator (URL) of a website or other services. By default, HTTP uses port 80 and HTTPS uses port 443, but a URL like http://www.test.com:8000/test1/ specifies that the web site is served by the HTTP server on port 8000. After collecting datasets the Apriori algorithm creates a model to detect activities (IP or port) which are malicious. Table 2 shows some rules with their support and confidence numbers that are generated for users' activities.

TABLE 2. Misuse detection vs. Anomaly detection

| Row | Activity | Support, Confidence |
|-----|----------|---------------------|
| 1 | 192.168.1.154:81 | [0.845154,0.15185] |
| 2 | 192.168.1.154:83 | [0.813043,0.02568] |
| 3 | 192.168.1.160 | [0.491373,0.24587] |
| 4 | 192.168.1.127:8043 | [0.298312,0.14548] |
| 5 | 192.168.1.114:8485 | [0.661538,0.05623] |

For example, in row 1 in table 2, Apriori algorithm creates a model that contains this fact:

192.168.1.154:81 => INTRUSION, [0. 845154, 0. 15185]

Meaning: 84.5154% of the time when user does ACTIVITY (192.168.1.154:83), INTRUSION (a kind of intrusive activity) is occurred; and the INTRUSION constitutes 15.185% of all detected intrusions.

This method generates frequent rules for firewall whose occurrences exceed a predefined minimum support threshold. The threshold in proposed model is equal 70% and in this example the support exceeds 70% therefore in the firewall this rule is generated.

## VIII. CONCLUSION

This paper proposed a model to improve anomaly intrusion detection by generating real-time rules for firewall. This model was proposed to use systematic data mining approaches to select the relevant system features to build better rules for firewall. For this purpose Apriori algorithm was used. This algorithm extracts interesting correlation relationships among large set of data items. Snort is used to record logs of user activities, and then Apriori algorithm will be used to create a model. This model can be used to create online rules for firewall based on current user activities.